\documentclass[11pt]{article}
\usepackage{coling2020}
\usepackage{times}
\usepackage{url}
\usepackage{latexsym}

\colingfinalcopy

\usepackage[utf8]{inputenc} 
\usepackage[T1]{fontenc}    
\usepackage{hyperref}       
\usepackage{url}            
\usepackage{booktabs}       
\usepackage{amsfonts}       
\usepackage{nicefrac}       
\usepackage{microtype}      
\usepackage{lipsum}
\usepackage{graphicx}
\usepackage{amsmath}
\usepackage{csquotes}
\usepackage{makecell}

\usepackage{microtype}
\usepackage{multirow}
\usepackage{latexsym}
\usepackage{booktabs}
\usepackage{courier}
\usepackage{amsmath}
\usepackage{placeins}
\usepackage{tikz}
\usepackage{hyperref}

\newcommand{\ra}[1]{\renewcommand{\arraystretch}{#1}}

\title{Scientific Explanation and Natural Language: \\A Unified Epistemological-Linguistic  Perspective for Explainable AI}

\author{
  Marco Valentino and Andr\'e Freitas\\
  Idiap Research Institute, Switzerland\\
  University of Manchester, United Kingdom\\
  \texttt{\{marco.valentino,andre.freitas\}@idiap.ch} \\
}

\begin{document}
\maketitle


\begin{abstract}
A fundamental research goal for Explainable AI (XAI) is to build models that are capable of reasoning through the generation of \emph{natural language explanations}. However, the methodologies to design and evaluate explanation-based inference models are still poorly informed by theoretical accounts on the nature of explanation. As an attempt to provide an epistemologically grounded characterisation for XAI, this paper focuses on the scientific domain, aiming to bridge the gap between theory and practice on the notion of a \emph{scientific explanation}. Specifically, the paper combines a detailed survey of the modern accounts of scientific explanation in Philosophy of Science with a systematic analysis of corpora of natural language explanations, clarifying the nature and function of explanatory arguments from both a top-down (categorical) and a bottom-up (corpus-based) perspective. 
Through a mixture of quantitative and qualitative methodologies,
the presented study allows deriving the following main conclusions: (1) Explanations cannot be entirely characterised in terms of \emph{inductive} or \emph{deductive} arguments as their main function is to perform \emph{unification}; (2) An explanation must cite \emph{causes} and \emph{mechanisms} that are responsible for the occurrence of the event to be explained; (3) While natural language explanations possess an intrinsic causal-mechanistic nature, they are not limited to causes and mechanisms, also accounting for pragmatic elements such as \emph{definitions}, \emph{properties} and \emph{
taxonomic relations}; (4) Patterns of \emph{unification} naturally emerge in corpora of explanations even if not intentionally modelled; (5) Unification is realised through a process of \emph{abstraction}, whose function is to provide the inference substrate for subsuming the event to be explained under recurring patterns and high-level regularities. The paper contributes to addressing a fundamental gap in classical theoretical accounts on the nature of scientific explanations and their materialisation as linguistic artefacts. This characterisation can support a more principled design and evaluation of explanation-based AI systems which can better interpret and generate natural language explanations. 
\end{abstract}

\section{Introduction}

Building models capable of performing complex inference through the generation of \emph{natural language explanations} represents a fundamental research goal for explainabilty in AI \cite{dovsilovic2018explainable,danilevsky2020survey,thayaparan2020survey}. 
However, while current lines of research focus on the development of explanation-based models and benchmarks \cite{wiegreffe2021teach,dalvi2021explaining,xie2020worldtree,jhamtani2020learning,jansen2018worldtree,thayaparan-etal-2021-textgraphs}, the applied methodologies are still poorly informed by formal accounts and discussions on the nature of explanation \cite{woodward2003scientific,miller2018explanation,tan2021diversity}. When describing natural language explanations, in fact, existing work rarely recur to formal characterisations of what constitute a \emph{valid explanatory argument}, and are often limited to the indication of generic properties in terms of \emph{supporting evidence} or \emph{entailment} \cite{yang2018hotpotqa,camburu2018snli,valentino2021natural,dalvi2021explaining}. Bridging the gap between between theory and practice, therefore, can accelerate progress in the field, providing new opportunities to formulate clearer research objectives and improve the evaluation methodologies \cite{camburu-etal-2020-make,valentino2021natural,jansen2021challenges,clinciu-etal-2021-study}.

As an attempt to provide an epistemologically grounded characterisation for Explainable AI (XAI), this paper aims to bridge the gap in the notion of \emph{scientific explanation} \cite{salmon2006four,salmon1984scientific}, studying it as both a \emph{formal object} and as a \emph{linguistic expression}. 

To this end, the paper is divided in two main sections.
The first part represents a systematic survey of the modern discussion in Philosophy of Science around the notion of a scientific explanation, shading light on the nature and function of explanatory arguments and their constituting elements \cite{hempel1948studies,kitcher1989explanatory}.
Following the survey, the second part of the paper presents a corpus analysis aimed at qualifying sentence-level \emph{explanatory patterns} in corpora of natural language explanations, focusing on datasets used to build and evaluate explanation-based inference models in the scientific domain \cite{xie2020worldtree,jansen2014discourse}. 

Overall, the paper presents the following main conclusions:

\begin{enumerate}
    \item \textbf{Explanations cannot be exclusively characterised in terms of \emph{inductive} or \emph{deductive} arguments.} Specifically, the main function of an explanation is not of \emph{predicting} or \emph{deducing} the event to be explained (\emph{explanandum}) \cite{hempel1965aspects}, but the one of showing how the explanandum fits into a  \emph{broader underlying regularity}. This process is known as \emph{unification}, and it is responsible for the creation of \emph{explanatory patterns} that can account for a large set of phenomena \cite{friedman1974explanation,kitcher1981explanatory}.
    \item \textbf{An explanation must cite part of the causal history of the explanandum}, fitting the event to be explained into a \emph{causal nexus} \cite{salmon1984scientific}. There are two possible ways of constructing causal explanations: (1) an explanation can be \emph{etiological} -- i.e., the explanandum is explained by revealing part of its causes -- or (2) \emph{constitutive} -- i.e., the explanation describes the underlying mechanism giving rise to the explanandum. This is confirmed by the corpus analysis, which reveals that the majority of natural language explanations, indeed, contain references to mechanisms and/or direct causal interactions between entities \cite{jansen2014discourse}.
    \item \textbf{While explanations possess an intrinsic causal-mechanistic nature, they are not limited to causes and mechanisms.} In particular, additional knowledge categories such as \emph{definitions}, \emph{properties} and \emph{taxonomic relations} seem to play an equally important role in building an explanatory argument. This can be attributed to both \emph{pragmatic aspects} of natural language explanations as well as inference requirements associated to \emph{unification}.
    
    \item \textbf{Patterns of unification naturally emerge in corpora of explanations.} Even if not intentionally modelled,  \emph{unification} seems to be an emergent property of corpora of natural language explanations \cite{xie2020worldtree}. The corpus analysis, in fact, reveals that the distribution of certain explanatory sentences is connected to the notion of \emph{unification power} and that it is possible to draw a parallel between inference patterns emerging in natural language explanations and formal accounts of explanatory unification \cite{kitcher1989explanatory}.
    
    \item \textbf{Unification is realised through a process of abstraction.} Specifically, abstraction represents the fundamental inference substrate supporting unification in natural language, connecting concrete instances in the explanandum to high-level concepts in central explanatory sentences. This process, realised through specific linguistic elements such as definitions and taxonomic relations, is a fundamental part of natural language explanations, and represents what allows subsuming the event to be explained under high-level patterns and unifying regularities.
\end{enumerate}

We conclude by suggesting how the presented findings can open new lines of research for explanation-based AI systems, informing the way the community should approach model creation and the design of evaluation methodologies for natural language explanations.

The paper contributes to addressing a fundamental gap in classical theoretical accounts on the nature of scientific explanations and their materialisation as linguistic artefacts. This characterisation can support a more principled design of AI systems that can better interpret and generate natural language explanations.
To the best of our knowledge, while previous work on natural language explanations have performed quantitative and qualitative studies in terms of knowledge reuse and inference categories \cite{jansen2016s,jansen2017study}, this study is the first to explore the relation between linguistic aspects of explanations and formal accounts in Philosophy of Science \cite{woodward2003scientific}, providing a unified epistemological-linguistic perspective for the field.

\section{Scientific Explanation: The Epistemological Perspective}

\begin{table*}[t]
\centering
\small
\ra{1}
\resizebox{\textwidth}{!}{
\begin{tabular}{p{5cm}p{5cm}p{5cm}}
\toprule
\textbf{Account} &
\textbf{Explanans} & \textbf{Relation}\\
\midrule
\textbf{Epistemic}\\
\midrule
Deductive-Nomological \cite{hempel1948studies} & Initial conditions + at least a universal law of nature & The \emph{explanandum} is logically deduced from the \emph{explanans}\\
\midrule
Inductive-Statistical \cite{hempel1965aspects} & Initial conditions + at least a statistical law & The \emph{explanans} make the \emph{explanandum} highly probable\\
\midrule
Unificationist \cite{kitcher1989explanatory} & A theory T + a class of phenomena P including the \emph{explanandum} & Shows how a class of phenomena P can be derived from a theory T through the instantiation of an argument pattern\\
\toprule
\textbf{Ontic}\\
\midrule
Statistical-Relevance \cite{salmon1971statistical} & A set of statistically relevant facts & the \emph{explanans} increase the probability of the \emph{explanandum}\\
\midrule
Causal-Mechanical \cite{salmon1984scientific} & A set of relevant causal processes and interactions & The \emph{explanans} are part of the causal history of the \emph{explanandum}; the \emph{explanans} are part of the mechanism underlying the \emph{explanandum}\\
\bottomrule
\end{tabular}}
\caption{The main modern accounts of scientific explanation in Philosophy of Science.}
\label{tab:explanation_models}
\end{table*}

The ultimate goal of science goes far beyond the pure prediction of the natural world. Science is constantly seeking a deeper understanding of observable phenomena and recurring patterns in nature by means of scientific theories and explanations. Most philosophers define an explanation as an answer to a \emph{``why''} question, aiming at identifying and describing the reason behind the occurrence and manifestation of particular events \cite{salmon1984scientific}. However, although the explanatory role of science is universally acknowledged, a formal definition of what constitutes and characterise a scientific explanation remains a complex subject. This is attested by the long history of the discussion in Philosophy of Science, which goes back at least to Ancient Greece \cite{hankinson2001cause}.
Nevertheless, relatively recent attempts at delivering a rigorous account of scientific explanation have produced a set of quasi-formal models that clarify to some extent the nature of the concept \cite{salmon2006four}. 

The modern view of scientific explanation has its root in the work of Carl Gustav Hempel and Paul Oppenheim, \emph{``Studies in the Logic of Explanation''} \cite{hempel1948studies}, whose publication in 1948 raised a heated debate in the Philosophy of Science community \cite{woodward2003scientific}. This section will survey the main accounts resulting from this debate with the aim of summarising and revisiting the main properties of a scientific explanation.
In particular, the goal of the survey is to identify the principal constraints that these models impose on \emph{explanatory arguments}, highlighting their essential features and function. This analysis will lead to the comprehension of the essential characteristics that differentiate explanation from other types of knowledge in science, such as prediction. 

In general, an explanation can be described as an argument composed of two main elements:
\begin{enumerate}
    \item The \emph{Explanandum}: the fact representing the observation or event to be explained.
    \item The \emph{Explanans}: the set of facts that are invoked and assembled to produce the explanation.
\end{enumerate}

The aim  of a formal account of a scientific explanation is to define an \emph{``objective relationship''} that connects the explanandum to the explanans  \cite{salmon1984scientific}, imposing constraints on the class of possible arguments that constitute a valid explanation. 
Existing accounts, therefore, can be classified according to the nature of the relationship between explanans and explanandum (Table \ref{tab:explanation_models}). Specifically, this survey will focus on accounts falling under two main conceptions:
\begin{itemize}
    \item \emph{Epistemic:} The explanation is an \emph{argument} showing how the explanandum \emph{can be derived} from the explanans. There is a relation of \emph{logical necessity} between the explanatory statements and the event to be explained. 
    \item \emph{Ontic:} The explanation relates the explanandum to \emph{antecedent conditions} by means of general laws, \emph{fitting} the explanandum into a \emph{discernible pattern}.
\end{itemize}


\subsection{Explanation as an Argument}

\begin{figure*}[t]
\centering
\includegraphics[width=\textwidth]{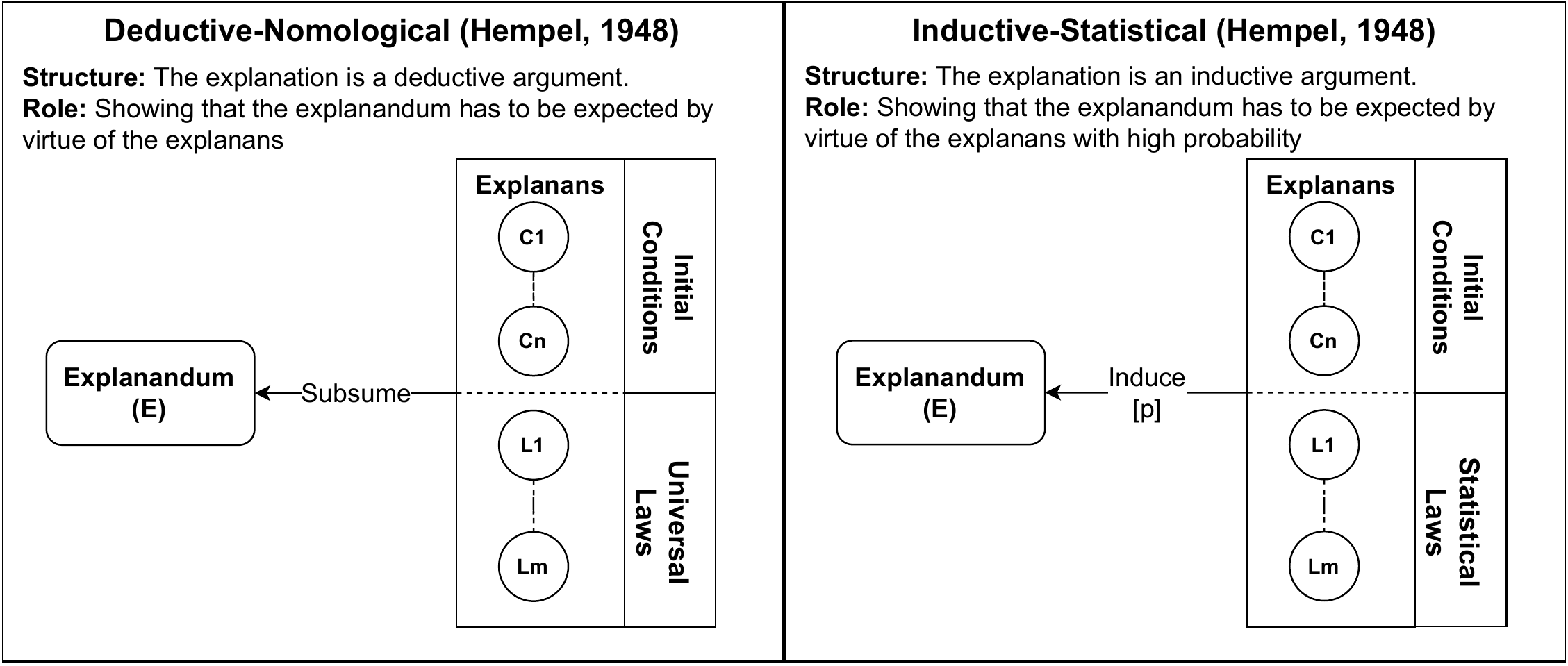}
\caption{Schematic representation of the Deductive-Inductive account of scientific explanation.}
\label{fig:epistemic}
\end{figure*}

\subsubsection{Deductive-Inductive Arguments}

The \emph{Deductive-Nomological (DN)} model proposed by Hempel \cite{hempel1948studies} is considered the first modern attempt to formally characterise the concept of scientific explanation. the DN account defines an explanation as an argument, connecting explanans and explanandum by means of \emph{logical deduction}.  Specifically, the explanans constitute the premises of a deductive argument while the explanandum represents its logical conclusion.
The general structure of the DN model can be schematised as follows:

$$
     C_I,C_2 \dots ,C_k \quad \text{Initial Conditions}
$$
$$
     \frac{L_I,L_2 \dots ,L_r \quad \text{Universal Laws of Nature} }{E \quad \text{Explanandum}}
$$

In this model, the explanans are constituted by a set of initial conditions, $C_1,C_2 \ldots, C_k$, plus at least a universal law of nature, $L_1,L_2 \ldots, L_r$ (with $r > 0$). According to Hempel, in order to represent a valid scientific argument, an explanation must include only explanans that are empirically testable. At the same time, the universal law must be a statemet describing a \emph{universal} regularity, while the initial conditions represent particular facts or phenomena that are concurrently observable with the explanandum.
Here is a concrete example of a scientific explanation under the DN account \cite{hempel1965aspects}:
\begin{itemize}
\item $C_1$: The (cool) sample of mercury was placed in hot water;
\item $C_2$: Mercury is a metal;
\item $L_1$: All metals expand when heated;
\item $E$: The sample of mercury expanded.
\end{itemize}

To complete the DN account with a theory of statistical explanation, Hempel introduced the \emph{Inductive-Statistical (IS)} model \cite{hempel1965aspects}. According to the IS account, an explanation must show that the explanandum was to be expected with \emph{high probability} given the explanans. Specifically, an explanation under the IS account has the same structure of the DN account, replacing the universal laws with statistical laws.
In order for a statistical explanation to be appropriate, the explanandum must be induced from statistical laws and initial conditions with probability close to 1.

The Deductive-Inductive view proposed by Hempel emphasises the \emph{predictive power} of explanations. Given a universal/statistical law and a set of initial conditions, it is possible to establish whether or not a particular phenomenon will occur in the future. According to Hempel, in fact, not only predictive power is a fundamental property of an explanation, explanations and predictions share exactly the \emph{same logical structure}. Specifically, the only difference between explanatory and predictive arguments is when they are formulated or requested: explanations are generally required for past phenomena, while predictions for events that have yet to occur. 

This feature of the Deductive-Inductive account is known as the \emph{symmetry thesis} \cite{hempel1965aspects} which has been largely criticised by other philosophers in the field \cite{salmon1984scientific,kitcher1989explanatory}. 
The symmetry thesis, in fact, leads to well-known objections and criticisms of Hempel's account. 
Consider the following example:
\begin{itemize}
    \item $C_1$: The elevation of the sun in the sky is $x$;
    \item $C_2$: The height of the flagpole is $y$;
    \item $L_1$: Laws of physics concerning the propagation of light;
    \item $L_2$: Geometric laws;
    \item $E$: The length of the shadow is $z$.
\end{itemize}
While the example above represents a reasonable explanatory argument, the DN account does not impose any constraint that prevents the interchanging of the explanandum with some of the initial conditions:
\begin{itemize}
    \item $C_1$: The elevation of the sun in the sky is $x$;
    \item $C_2$: The length of the shadow is $z$;
    \item $L_1$: Laws of physics concerning the propagation of light;
    \item $L_2$: Geometric laws;
    \item $E$: The height of the flagpole is $y$.
\end{itemize}
The DN model and its symmetry property, in particular, allows for the construction of explanatory arguments that contain inverted causal relations between its elements.
This counterexample shows that prediction and explanation \emph{must have a different logical structure} and treated as different types of arguments.  Although predictive power is a necessary property of an adequate explanation, it is not sufficient. Explanations, in fact, are inherently \emph{asymmetric}, a property that cannot be described by means of deductive-inductive arguments alone.

In Hempel's account, moreover, there is a further property of explanation that has been subject to criticisms by subsequent philosophers, that is the notion of \emph{explanatory relevance}. Consider the following counter-example from Salmon \shortcite{salmon1984scientific}:
\begin{itemize}
    \item $C_1$: John Jones is a male;
    \item $C_2$: John Jones has been taking birth control pills regularly;
    \item $L_1$: Males who take birth control pills regularly fail to get pregnant;
    \item $E$: John Jones fails to get pregnant.
\end{itemize}
Although the argument is formally correct, it contains statements that are explanatorily irrelevant to $E$. Specifically, the fact that \emph{John Jones has taken birth control pills} should not be cited in an explanation for  \emph{John Jones fails to get pregnant}. In this particular example only $C_1$ is relevant to $E$, and only $C_1$ should figure into an explanation for $E$. Specifically, the universality and high probability requirements of the DN and IS model constrain the explanation to include all the explanatory relevant premises but not to exclude irrelevant facts \cite{salmon1984scientific}.

\begin{figure}[t]
\centering
\includegraphics[width=0.5\textwidth]{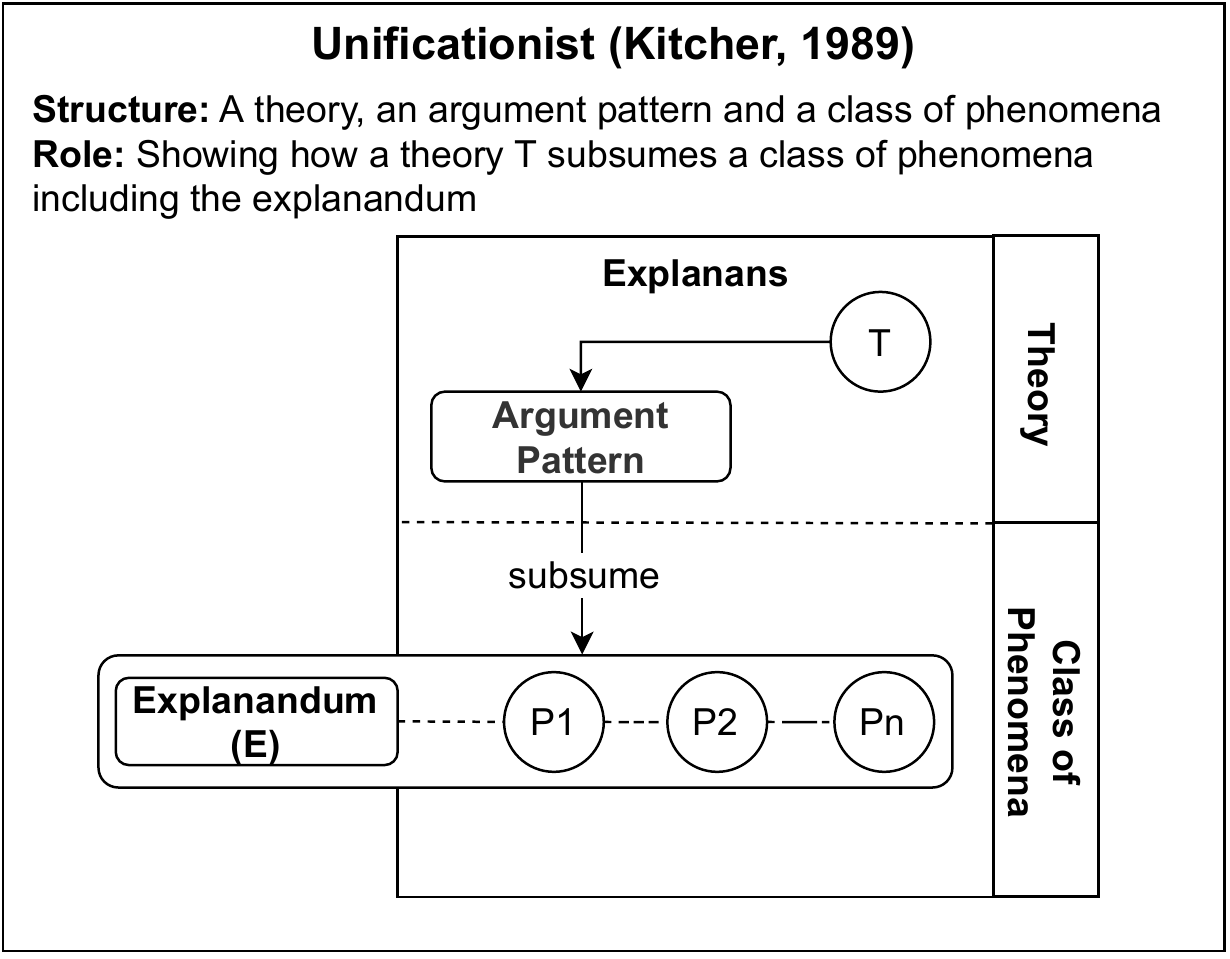}
\caption{A schematic representation of the Unificationist account of scientific explanation.}
\label{fig:explanatory_unification}
\end{figure}

\subsubsection{Explanatory Unification and Argument Patterns}

The Unificationist account of scientific explanation was proposed by Friedman \cite{friedman1974explanation} and subsequently refined by Kitcher \cite{kitcher1989explanatory,kitcher1981explanatory} in order to overcome the criticisms, including relevance and asymmetry, raised by the Deductive-Inductive account. 

According to the Unificationist model, an explanation cannot be uniquely described in terms of deductive or inductive arguments. To properly characterise an explanation, in fact, it is necessary to consider its main function of fitting the explanandum into a \emph{broader unifying pattern}. Specifically, an explanation is an argument whose role is to connect a set of \emph{apparently unrelated phenomena}, showing that they can be subsumed under a common underlying regularity. The concept of explanatory unification is directly related to the goal of Science of understanding nature by reducing the number of disconnected phenomena and provide an ordered and clear picture of the world \cite{schurz1999explanation}.

Figure \ref{fig:explanatory_unification} shows a schematic representation of the Unificationist account. Given a scientific theory $T$ and a class of phenomena $P$ including the explanandum $E$, an explanation is an argument that allows deriving all the phenomena in $P$ from $T$. In this case, we say that $T$ \emph{unifies} the explanandum $E$ with the other phenomena in $P$. According to Kitcher, a scientific explanation accomplishes unification by deriving descriptions of many phenomena through the same patterns of derivation \cite{kitcher1989explanatory}. Specifically, a theory defines an \emph{argument pattern} which can be occasionally instantiated to explain particular phenomena or observations. 

An argument pattern is a sequence of \emph{schematic sentences} organised in premises and conclusions. In particular, a schematic sentence can be described as a template obtained by replacing some non-logical expressions in a sentence with \emph{variables} or \emph{dummy letters}. For instance, from the statement  \emph{``Organisms homozygous for the sickling allele develop sickle-cell anemia''} it is possible to generate schematic sentences at different levels of abstraction: \emph{``Organisms homozygous for A develop P''} and \emph{``For all x, if x is O and A then x is P''}. An argument pattern can be instantiated by specifying a set of 
\emph{filling instructions} for replacing the variables of the schematic sentences together with rules of inference for the derivation. For example, a possible filling instruction for the schematic sentence \emph{``Organisms homozygous for A develop P''} might specify that $A$ must be substituted by the name of an allele and $P$ by some phenotypic trait. Different theories can induce different argument patterns whose structure is not defined a-priori as in the case of Hempel's account. However, once a theory is accepted, the same argument pattern can be instantiated to explain a large variety of phenomena depending on the unification power of the theory.

The history of science is full of theories and explanations performing unification, and the advancement of science itself can be seen as a process of growing unification \cite{friedman1974explanation}. A famous example is provided by Newton's law of universal gravitation, which unifies the motion of celestial bodies and falling objects on Earth showing that they are all manifestation of the same underlying physical law. Specifically, from the unificationist point of view, Newton's law of universal gravitation defines an argument pattern whose filling instructions apply to all entities with mass.

The Unificationist account provides a set of criteria to identify the \emph{``best explanation''} among competing theories: 
\begin{enumerate}
    \item \emph{Unification power}: Given a set of phenomena $P$ and a theory $T$. the larger is the cardinality of $P$ - i.e. the number of phenomena that are unified by $T$, the greater is the explanatory power of $T$.
 \item\emph{Simplicity}: Given two theories $T$ and $T_1$ able to unify the same set of phenomena $P$, the theory that makes use of a lower number of premises in its argument patterns is the one with the greatest explanatory power.
\end{enumerate}
These selection criteria play a fundamental role in the Unificationist account since, according to Kitcher, only the best explanation available at a given point in time should be considered as the valid one \cite{kitcher1981explanatory}. For example, to explain the motion of celestial bodies by means of gravity, one must consider Einstein's theory of relativity as the valid explanation, as it allows to subsume a broader set of phenomena compared to Newton's law of universal gravitation. 

The simplicity criteria prevents the explanation to include irrelevant premises as in the case of the control pill example analysed under the Deductive-Inductive account since, under the same unification power, an explanation containing less premises will be preferred over a more complex explanation introducing unnecessary statements. Similarly, the problem of asymmetry can be solved considering the unification power criteria. Specifically, argument patterns containing inverted causal relations will generally allow for the derivation of fewer phenomena. According to Kitcher, in fact, causality is an emergent property of unification:
\emph{``to explain is to fit the phenomena into a unified picture insofar as we can. What
emerges in the limit of this process is nothing less than the causal structure of the world''} \cite{kitcher1989explanatory}.

\subsection{Fitting the Explanandum into a Discernible Pattern}

\begin{figure*}[t]
\centering
\includegraphics[width=\textwidth]{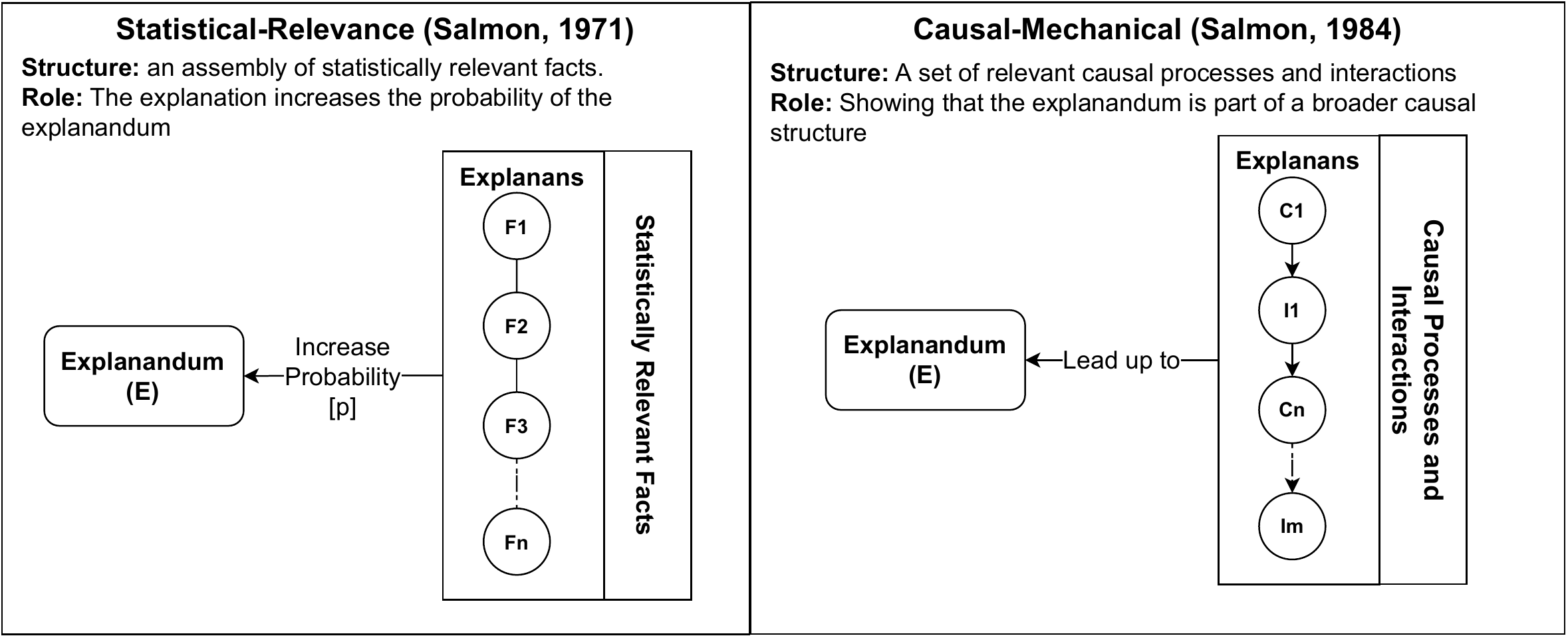}
\caption{Schematic representation of accounts falling under the \emph{ontic} conception.}
\label{fig:ontic}
\end{figure*}

\subsubsection{Statistical-Relevance}

Motivated by the problem of relevance in the Deductive-Inductive account, Wesley Salmon elaborated a statistical account of explanation known as \emph{Statistical Relevance (SR)} \cite{salmon1971statistical}. Differently from the Deductive-Inductive account, the SR model does not concern with the general structure and organisation of the explanatory argument, but attempts to characterise a scientific explanation in terms of the intrinsic relation between each explanatory statement and the explanandum.

In general, given a population $A$, a factor $C$ and some event $B$, we say that $C$ is \emph{statistically relevant} to the occurrence of $B$ if and only if 
\begin{equation}
    \small
    P(B|A.C) \neq P(B|A) \vee P(B|A.C) \neq P(B|A.\neg C)
\end{equation}
In other words, a given factor C is statistically relevant to an event B if its occurrence changes the probability of B to occur. According to the SR account, the explanatory relevance of a fact has to be defined in terms of its statistical relevance. Specifically, an explanation is an \emph{assembly of statistically relevant facts} that increase the probability of the explanandum. 

Consider the birth control pills example analysed under the IS account:
\begin{itemize}
    \item $C_1$: John Jones is a male;
    \item $C_2$: John Jones has been taking birth control pills regularly;
    \item $E$: John Jones fails to get pregnant.
\end{itemize}
Given a population $T$, we can perform a statistical analysis to verify whether $C_1$ and $C_2$ are relevant to E:
\begin{equation}
\small
    P(pregnant|T.male) = P(pregnant|T.male.pills)
\end{equation}
\begin{equation}
\small
    P(pregnant|T.pills) \neq P(pregnant|T.pills.male)
\end{equation}

Notice that in (2.2), given the fact that a generic $x \in T$ is a male ($T.male$), the action of taking birth control pills ($T.male.pills$) has no affect on the probability that $x$ is pregnant. Conversely, in (2.3), the probability that a generic member of the population $x$ is pregnant, given the action of taking pills ($T.pills$), decreases to zero if we know that $x$ is a male ($T.pills.male$). Therefore, the statistical relevance analysis leads to the conclusion that  \emph{``among males, taking birth control pills is explanatorily irrelevant to pregnancy, while being male is relevant''} \cite{salmon1984scientific}.

The SR model shows that a fact can be explanatorily relevant even if it does not induce the explanandum with a probability close to 1. Specifically, the relevance depends on the effect that the explanans have on the probability of the explanandum rather than on its absolute value.  Contrary to the Inductive-Statistical account, this property guarantees the possibility to formulate explanations for rare phenomena.

\begin{figure}[t]
\centering
\includegraphics[width=0.4\textwidth]{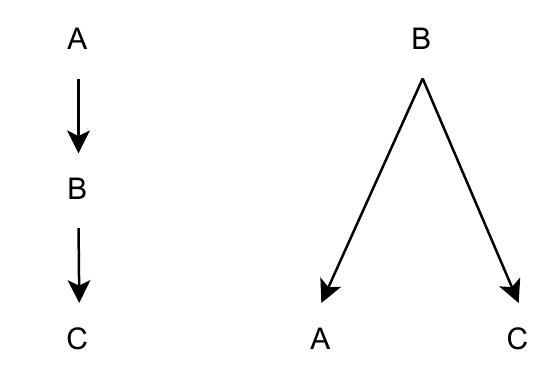}
\caption{Causal relationships are underdetermined by statistical relevance relationships. In this example, in particular, it is not possible to discriminate between the depicted causal structures using a statistical relevance analysis. In both cases, in fact, $A$ is statistically relevant to $C$; a factor that can lead, in the situation depicted on the right, to a SR explanation based on the relation between $A$ and $C$ induced by the common cause $B$.}
\label{fig:sr_causal_structures}
\end{figure}

Although statistical relevance seemed to provide a formal way to shield explanation from irrelevance, Salmon subsequently realised that the SR model is not sufficient to elaborate an adequate account of scientific explanation \cite{salmon1984scientific,salmon1998causality}. It is nowadays clear, in fact, that certain causal structures are greatly underdetermined by statistical relevance \cite{pearl2009causality,pearl2019seven}. Specifically, different causal structures can be described by the same statistical relevance relationships among their elements, making it impossible to discriminate direct causal links by means of statistical relevance analysis alone (Figure \ref{fig:sr_causal_structures}).

 According to Salmon, \emph{``the statistical relationships specified in the SR model constitute the statistical basis for a bona-fide scientific explanation, but this basis must be supplemented by certain causal factors in order to constitute a satisfactory scientific explanation''} \cite{salmon1984scientific}. The failed attempt to characterise a scientific explanation uniquely in terms of statistical elements demonstrated, as in the case of Hempel's account, the intrinsic difference between prediction and explanation. The latter, in fact, cannot be derived by pure statistical observations and seems to require conjectures and hypotheses about hidden structures, such as the one induced by causal relations and interactions.

\subsubsection{Causes and Mechanisms}

Following the observation that the SR model is not sufficient for characterising a scientific explanation, Salmon formulated a new account known as the Causal-Mechanical (CM) model, in which the role of an explanation is to show how the explanandum fits into the \emph{causal structure of the world}. 
 Specifically, a valid scientific explanation cannot be limited to statistical relevance and must \emph{cite part of the causal history} leading up to the explanandum.


To formalise the CM account, Salmon attempted to define a theory of causality based on the concepts of \emph{causal processes} and \emph{interactions} \cite{salmon1998causality}. 
Consider the following example from \cite{woodward2005making}: 
\emph{``a cue ball, set in motion by the impact of a cue stick, strikes a stationary 8 ball with the result that the 8 ball is put in motion and the cue ball changes direction''}.
Here, the cue ball, the cue stick and the 8 ball are \emph{causal processes} while the collisions between the objects are \emph{causal interactions}. According to the CM model, the motion of the 8 ball has to be explained in terms of the causal processes and interactions belonging to its causal history. Therefore, a generic event $X$ is explanatorily relevant to the explanandum $E$ if and only if the following conditions apply:
\begin{enumerate}
    \item $X$ is statistically relevant to $E$
    \item $X$ and $E$ are part of different causal processes
    \item There exists a sequence of causal processes and interactions between X and E leading up to E
\end{enumerate}


Salmon identifies two major ways of constructing causal explanations for some event E. An explanation can be either \emph{etiological} -- i.e. E is explained by revealing part of its causes -- or \emph{constitutive} -- i.e. the explanation of E describes the underlying mechanism giving rise to E. 
A mechanism, in particular, is often described as an organised set of entities and activities, whose interaction is responsible for the emergence of a phenomenon \cite{craver2015mechanisms,craver2007top}. 
For example, it is possible to formulate an etiological explanation of a certain disease by appealing to a particular virus, or provide a constitutive explanation describing the underlying mechanisms by which the virus causes the disease.

The foremost merit of the CM account is to exhibit the profound connection between causality, mechanisms, and explanation, highlighting how most of the fundamental characteristics of a scientific explanation derive from its causal nature. Moreover, the differentiation between etiological and constitutive explanation had a significant impact on several scientific fields. Discovering mechanistic explanations, in fact, is nowadays acknowledged as the ultimate goal of many scientific disciplines such as biology and natural sciences \cite{craver2013search,schickore2014scientific,craver2015mechanisms,bechtel2005explanation}.

The CM model is still subject to a number of criticisms concerning the concepts of causal processes and interactions, which has led subsequent philosophers  to propose new theories of causality \cite{lewis1986causal,woodward2005making,hitchcock1995salmon}. However, the causal nature of scientific explanations is largely accepted, with much of the contemporary discussion focusing on philosophical and metaphysical aspects concerning causes and effects \cite{pearl2009causality}.

\begin{table*}
\centering
\small
\ra{1}
\resizebox{\textwidth}{!}{
\begin{tabular}{p{5cm}p{5cm}p{5cm}}
\toprule
 \textbf{Type of implied question} & \textbf{Type of contrast case} & \textbf{Type of cause}\\
\midrule
 \emph{``Why X rather than not X?''} & Non occurrence of effect & Sum of necessary conditions\\
 \midrule
 \emph{``Why X rather than the
default value for X?''} & The normal
case & Abnormal
condition\\
\midrule
\emph{``Why X rather than Y?''} & Noncommon
effect & Differentiating factor\\
\midrule
\emph{``Why X rather than what ought to be the case?''} & Prescribed or
statutory case & Moral or legal fault\\
\midrule
\emph{``Why X rather than the ideal
value for X?''} & Ideal case & Design fault or bug\\
\bottomrule
\end{tabular}}
\caption{Models of causal attribution adopted to answer different causal questions as defined by varying contrast cases (Hilton, 1990).}
\label{tab:contrastive_causal_explanation}
\end{table*}

An additional criticism is related to the inherent incompleteness of causal explanations \cite{hesslow1988problem,craver2020more}. Since the causes of some event can be traced back indefinitely, causal explanations must show only part of the causal history of the explanandum. This implies that not all the causes of an event can be included in an explanation. In Salmon's account, however, it is not clear what are the criteria that guide the inclusion of relevant causes and the exclusion of others. Subsequent philosophers claimed that the problem of relevance is context-dependent and that it can be only addressed by looking at explanations from a pragmatic perspective \cite{van1980scientific}. All why questions, in fact, seem to be \emph{contrastive} in nature \cite{lipton1990contrastive,miller2018contrastive}. Specifically, once a causal model is known, the explanans selected for a particular explanation depend on the specific why question, including only those causes that \emph{make the difference} between the occurrence of the explanandum and some \emph{contrast case} implied by the question \cite{miller2018explanation,hilton1990conversational} (Table \ref{tab:contrastive_causal_explanation}).

\subsection{Summary}

This section presented a detailed overview of the main modern accounts of scientific explanation, discussing their properties and limitations. 

Despite a number of open questions remain in the Philosophy of Science community, it is possible to draw the following conclusions:
\begin{enumerate}
    \item \textbf{Explanations and predictions have a different structure.} Any attempt to characterise a scientific explanation uniquely in terms of predictive elements has encountered fundamental issues from both an epistemic and an ontic perspective. An explanation, in fact, cannot be entirely characterised in terms of \emph{deductive-inductive arguments} or \emph{statistical relevance} relationships. This because predictive power, despite being a necessary property of a scientific explanation, is not a sufficient one.
    
    \item \textbf{Explanatory arguments create unification.} From an epistemic perspective, the main function of an explanatory argument is to fit the explanandum into a \emph{broader unifying pattern}. Specifically, an explanation must connect a class of \emph{apparently unrelated phenomena}, showing that they can be subsumed under a common underlying regularity. Form a linguistic point of view, the unifying power of explanations produces \emph{argument patterns}, whose instantiation can be used to explain a large variety of phenomena through the same patterns of derivation. 
    
    \item \textbf{Explanations possess an intrinsic causal-mechanistic nature.} From an ontic perspective, a scientific explanation must cite part of the causal history of the explanandum, fitting the event to be explained into a \emph{causal nexus}. There are two possible ways of constructing causal explanations: (1) an explanation can be \emph{etiological} -- i.e., the explanandum is explained by revealing part of its causes -- or (2) \emph{constitutive} -- i.e., the explanation describes the underlying mechanism giving rise to the explanandum.
\end{enumerate}

Philosophers tend to agree that the causal and unificationist accounts are complementary to each other, advocating for a \emph{``peaceful coexistence''} and a pluralistic view of scientific explanation \cite{salmon2006four,woodward2003scientific,strevens2004causal,bangu2017scientific,glennan2002rethinking}.
Unification, in fact, seems to be an essential property of causal explanations since many physical processes are the result of the same underlying
causal mechanisms \cite{salmon1998causality,salmon2006four}. At the same time, the unifying power of constitutive explanations derives from the existence of mechanisms that have a common higher-level structure, despite differences in the specific entities composing them \cite{glennan2002rethinking}.

Moreover, the unificationist account seems to be connected with theories of explanation and understanding in cognitive science, which emphasise the relationship between the process of searching for broader regularities and patterns to the way humans construct explanations in everyday life through abductive reasoning, abstraction, and analogies \cite{lombrozo2006structure,lombrozo2012explanation,keil2006explanation}.

\section{Scientific Explanation: The Linguistic Perspective}

The previous section focused on the notion of a scientific explanation from a quasi-formal (categorical) perspective, reviewing the main epistemological accounts attempting to characterise the space of valid explanatory arguments. Following this survey, this section assumes a linguistic perspective, investigating how the main features of the accepted accounts manifest in \emph{natural language}.

\begin{table}[t]
\small
 \centering
 \begin{tabular}{l|cc}
      \toprule
        \textbf{Feature} &
        \textbf{Why Corpus} &\textbf{WorldTree}\\
         \midrule
            Size & 193 & 2206\\
            Domain & Biology & Science exams\\
            Type & Scientific & Scientific - Commonsense\\
            Annotation & Textbooks & Manually curated\\
            Structured & No & Yes\\
            Reuse & No & Yes\\
    \bottomrule
 \end{tabular}
 \caption{Main features of the analysed explanations corpora. }
 \label{tab:scientific_explanation_corpora_features}
\end{table}

To this end, we present a systematic analysis of corpora of scientific explanations in natural language adopting a mixture of qualitative and quantitative methodologies to investigate the emergence of \emph{explanatory patterns} at both \emph{sentence} and \emph{inter-sentence} level, relating them to the \emph{Causal-Mechanical} \cite{salmon1998causality} and \emph{Unificationist} account \cite{kitcher1981explanatory,kitcher1989explanatory}. Specifically, we hypothesise that it is possible to map linguistic aspects emerging in natural language explanations to the discussed models of scientific explanation. At the same time, we observe that some linguistic and pragmatic elements in natural language explanations are not considered by the epistemological accounts, and therefore expect the corpus analysis to provide complementary insights on the nature of explanations as manifested in natural language. Bridging the gap between these two domains aims to provide a necessary linguistic-epistemological grounding for the construction of explanation-based AI models.

\begin{table*}[t]
\centering
\small
\ra{1}
\resizebox{\textwidth}{!}{
\begin{tabular}{p{6cm}p{7cm}p{4cm}}
\toprule
\textbf{Explanandum} & \textbf{Explanans} & \textbf{Knowledge Type}\\
\midrule
It is important for blood transfusions to not occur between individuals with different blood types & Certain bacteria normally present in the body have epitopes very similar to the A and B carbohydrates & Analogy, Comparison \\
\midrule
It is important for blood transfusions to not occur between individuals with different blood types & By responding to the bacterial epitope similar to the B carbohydrate, a person with type A blood makes antibodies that will react with the type B carbohydrate & Process, Mechanism \\
\midrule
It is important for blood transfusions to not occur between individuals with different blood types & Matching compatible blood groups is critical for safe blood transfusions & Requirement, Constraint \\
\midrule
Inbreeding does not cause evolution directly & The Hardy-Weinberg is a principle that describes a hypothetical population that is not evolving & Definition\\
\midrule
Inbreeding does not cause evolution directly & The gene pool is modified if mutations alter alleles or if entire genes are deleted or duplicated & Conditional, If-then \\
\midrule
Inbreeding does not cause evolution directly & Both inbreeding and genetic drift can cause a loss of genetic variation & Causal Interaction\\
\midrule
Inbreeding does not cause evolution directly & The allele and genotype frequencies often do change over time & Property, Attribute\\
\midrule
Steroids can easily pass through cell membranes & These complexes of a lipid-soluble hormone and its receptor act in the nucleus to regulate transcription of specific genes & Function, Roles\\
\midrule
Chromatin is important in meiosis & For example, the nuclei of human somatic cells (all body cells except the reproductive cells) each contain 46 chromosomes & Instances, Examples\\
\midrule
It is important for polypeptides to be able to greatly vary in amino acid sequence & Recall that most enzymes are proteins & Taxonimic, Meronymic\\
\midrule
Two traits that are more than 50cM away from each other are inherited randomly relative to each other & The observed frequency of recombination in crosses involving two such genes can have a maximum value of 50\% & Probability, Statistical Relevance\\
\bottomrule
\end{tabular}}
\caption{Explanation sentences in the Biology Why Corpus.}
\label{tab:biology_explanation_examples}
\end{table*}

\begin{table}[t]
\small
 \centering
 \begin{tabular}{p{15.5cm}}
      \toprule
        \textbf{Explanandum} \\
        \midrule
         Two sticks getting warm when rubbed together is
         an example of a force producing heat.\\
         \midrule
         \textbf{Explanans}\\
         \midrule
         (1) A stick is a kind of object; 
         (2) To rub together means to move against; 
         (3) Friction is a kind of force; 
         (4) Friction occurs when two object’s surfaces move against each other; 
         (5) Friction causes the temperature of an object to increase.\\
    \bottomrule
 \end{tabular}
 \caption{Example of a curated explanation in WorldTree.}
 \label{tab:worldtree_example}
\end{table}

The presented analysis focuses on two distinct corpora of explanations; the \emph{Biology Why Corpus} \footnote{https://allenai.org/data/biology-how-why-corpus} \cite{jansen2014discourse}, a dataset of biology why-questions with one or more explanatory passages identified in an undergraduate textbook, and the \emph{WordlTree Corpus} \footnote{http://cognitiveai.org/explanationbank/} \cite{xie2020worldtree}, a corpus of science exams questions curated with natural language explanations supporting the correct answers. 

The main features of the selected corpora are summarised in Table \ref{tab:scientific_explanation_corpora_features}. As shown in the table, the corpora have complementary characteristics. The explanations included in the \emph{Biology Why Corpus} are specific to a scientific domain (biology in this case), while the \emph{WorldTree Corpus} expresses a more diverse set of topics, including physics, biology, and geology. Since the explanatory passages from the \emph{Biology Why Corpus} are extracted from textbooks, the explanations tend to be more technical and unstructured. On the other hand, the explanations in \emph{WorldTree} are manually curated and represented in a semi-structured format (aiming more closely on inference automation), often integrating scientific sentences with commonsense knowledge. Moreover, the individual explanatory sentences in \emph{WorldTree} are reused across different science questions when possible, facilitating a quantitative study on knowledge use and the emergence of sentence-level explanatory patterns \cite{jansen2017study}. 

\begin{figure}[t]
\centering
\includegraphics[width=0.75\textwidth]{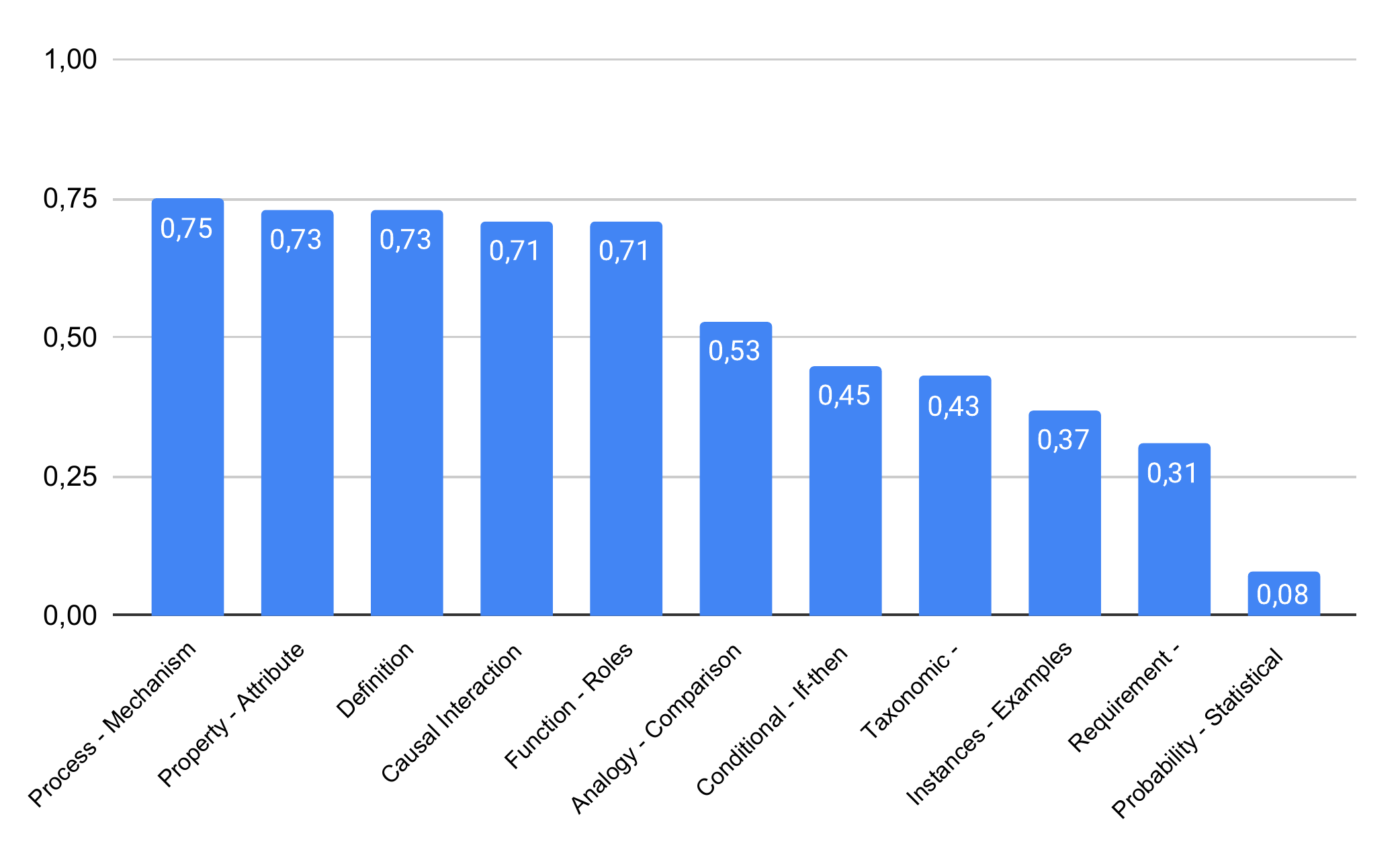}
\caption{Recurring knowledge in biological explanations.}
\label{fig:why_biology_corpus_analysis}
\end{figure}

By leveraging the complementary characteristics of the selected corpora and relating the corpus analysis to the discussed accounts of scientific explanation, we aim at investigating the following research questions:
\begin{enumerate}
    \item \textbf{RQ1:} What kinds of explanatory sentences occur in natural language explanations?
    \item \textbf{RQ2:} How do explanatory patterns emerge in natural language explanations?
\end{enumerate}

We adopt the \emph{Biology Why Corpus} and \emph{WorldTree} to investigate \textbf{RQ1}, while \emph{WorldTree} is considered for \textbf{RQ2} thanks to its size and reuse-oriented design.

\subsection{Biology Why Questions}

To study and investigate the emergence of sentence-level explanatory patterns in biological explanations we performed a systematic annotation of the explanatory passages included in the \emph{Biology Why Corpus} \cite{jansen2014discourse}. To this end, we identified a set of 11 recurring knowledge categories, annotating a sample of 50 explanations extracted from the corpus. Examples of annotated explanation sentences and their respective knowledge types are included in Table \ref{tab:biology_explanation_examples}. 


\subsubsection{Recurring Explanatory Sentences}

Figure \ref{fig:why_biology_corpus_analysis} reports the frequencies of each knowledge category in the annotated why-questions. Specifically, we consider each knowledge type as a binary variable (1 if the knowledge type appears in an answer to a why question, 0 otherwise) and compute a binomial distribution for each type.

The corpus analysis reveals that the majority of the why questions (75\%) are answered through the direct description of \emph{processes} and \emph{mechanisms}. As expected, this result confirms the crucial role of \emph{constitutive explanations} as  defined in the Causal-Mechanical (CM) account \cite{salmon1984scientific}. The importance of causality is confirmed by the frequency of sentences describing direct \emph{causal interactions} between entities (71\%), which demonstrates the interplay between \emph{constitutive} and \emph{etiological} explanation. Moreover, the analysis suggests that a large part of the explanations (71\%) include sentences describing \emph{functions} and \emph{roles}. The relation between the notion of function and mechanisms is reported in many constitutive accounts of explanation \cite{craver2015mechanisms}, and is typically understood as a mean of describing and situating some lower-level part within a higher-level mechanism \cite{craver2001role}.

The corpus analysis suggests that natural language explanations are not limited to causes and mechanisms and tend to include additional types of knowledge not explicitly discussed in the epistemological accounts. Specifically, the graph reveals that \emph{definitions} and sentences about \emph{attributes} and \emph{properties} play an equally important role in the explanations (both occurring in 73\% of the why questions). We attribute this result to \emph{pragmatic aspects} and inference requirements associated to the \emph{unification} process. 
Definitions, for instance, might serve both as a way to introduce missing context and background knowledge in natural language discourse and, in parallel, as a mechanism for \emph{abstraction}, relating specific terms to high-level conceptual categories \cite{silva2018recognizing,silva2019exploring,stepanjans2019identifying}.

The role of abstraction in the explanations is supported by the presence of \emph{analogies} and \emph{comparison} between entities (53\%), as well as sentences describing \emph{taxonomic} or \emph{meronymic} relations (43\%). These characteristics suggest the presence of explanatory arguments performing unification through an abstractive inference process, whose function is to identify common abstract features between concrete instances in the explanandum \cite{kitcher1981explanatory}. The role of abstraction will be explored in details in the next section.

Finally, the corpus analysis reveals a  low frequency of sentences describing \emph{statistical relevance} relationships and \emph{probabilities} (8\%). These results reinforce the fundamental difference between explanatory and predictive arguments identified and discussed in the philosophical accounts \cite{woodward2003scientific}.

\subsection{Science Questions}

\begin{figure}[t]
\centering
\includegraphics[width=0.65\textwidth]{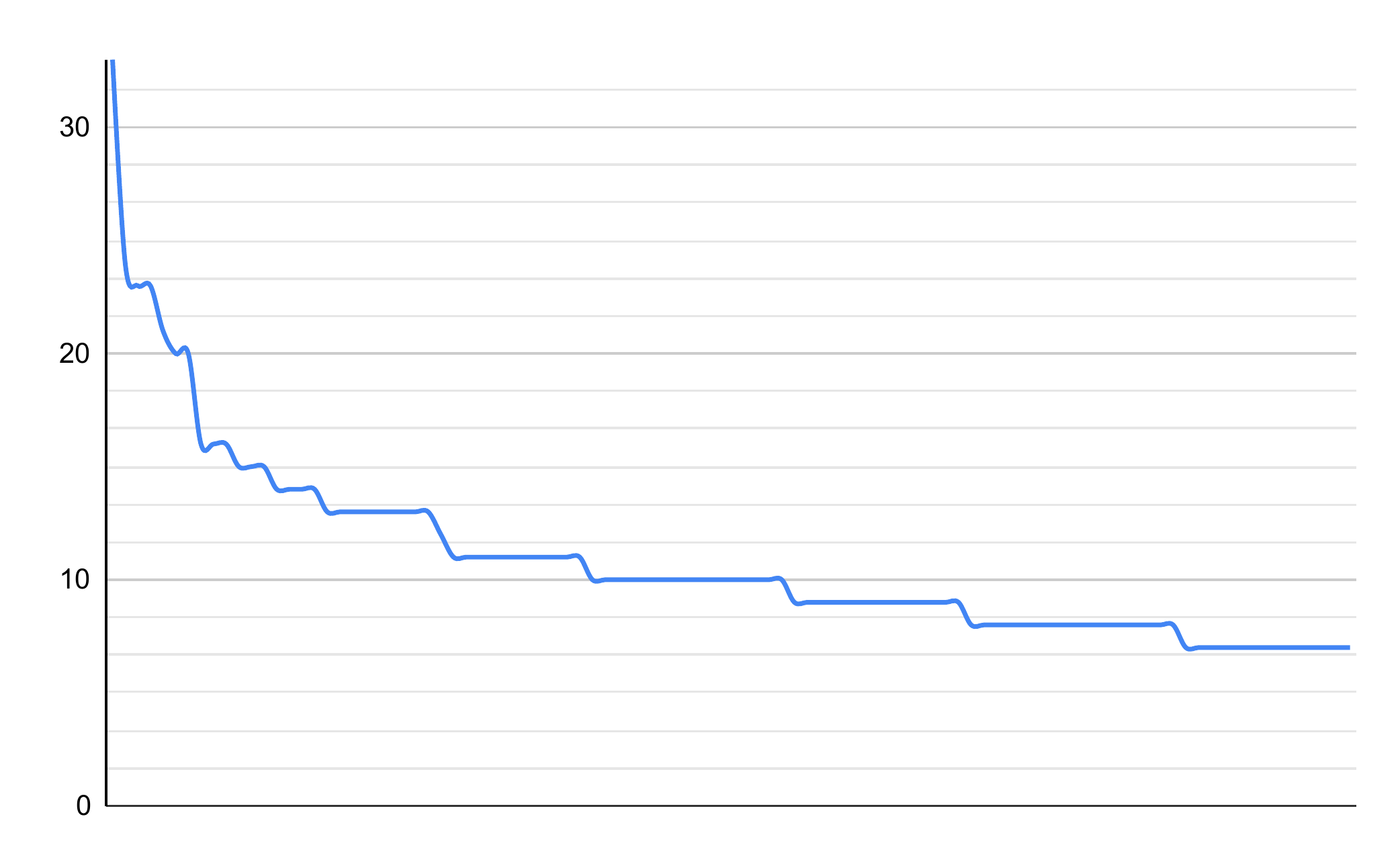}
\caption{Distribution and reuse of central explanatory sentences in WorldTree.}
\label{fig:central_dist}
\end{figure}

\begin{figure}[t]
\centering
\includegraphics[width=0.65\textwidth]{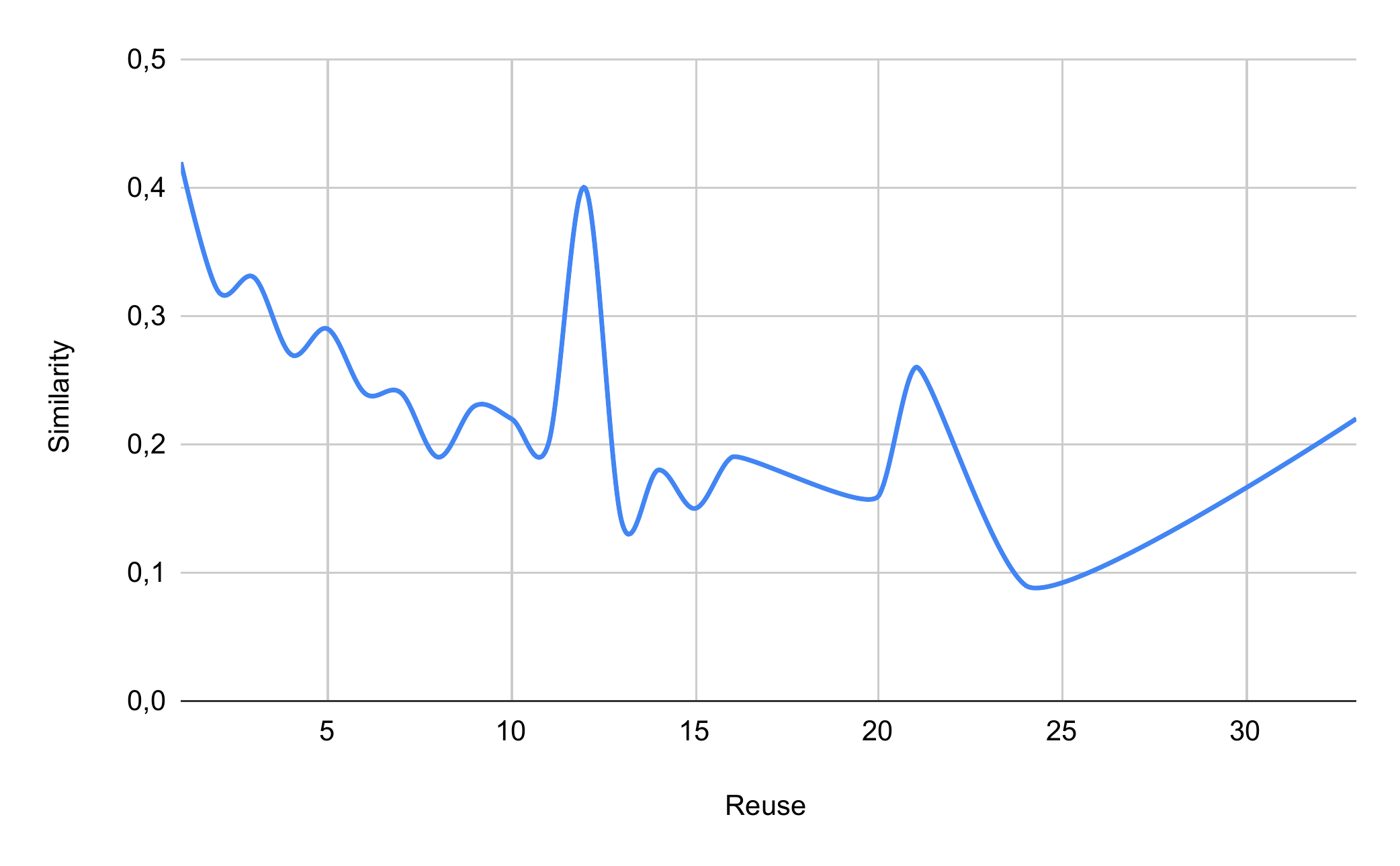}
\caption{Similarity between central sentences and questions vs frequency of reuse of the central sentences.}
\label{fig:central_occurrence_similarity}
\end{figure}


This section presents a corpus analysis on WorldTree \cite{xie2020worldtree} aimed at investigating the emergence of explanatory patterns and unification, relating them to epistemological aspects of scientific explanations. 
Table \ref{tab:worldtree_example} shows an archetypal example of explanation in WorldTree. Here, the explanandum is represented by a statement derived from a science question and its correct answer, while the explanans are an assembly of sentences retrieved from a background knowledge base. 

The corpus categorises the core explanans according to different explanatory roles:

\begin{itemize}
    \item \emph{Central:} Sentences explaining the central concepts that the question is testing.
    \item \emph{Grounding:} Sentences linking generic
    terms in a central sentence with specific instances of
    those terms in the question.
\end{itemize}

Some explanatory sentences in WorldTree can be categorised according to additional roles that are not strictly required for the inference (i.e., \emph{Background} and \emph{Lexical Glue} \cite{jansen2018worldtree}) and that, for the purpose of investigating the nature of explanatory patterns, will not be considered in the corpus analysis.

\subsubsection{Distribution and Reuse of Explanatory Sentences}


The first analysis concentrates on the distribution and reuse of \emph{central} explanatory sentences in the corpus. The quantitative results of this analysis are presented in Figure \ref{fig:central_dist} and \ref{fig:central_occurrence_similarity}, while a set of qualitative examples are reported in Table \ref{tab:central_facts_example}.

The graph in Figure \ref{fig:central_dist} describes the distribution of individual sentences annotated as central explanatory facts across different explanations. Specifically, the y-axis represents the number of times a specific sentence is used as a central explanation for a specific science question. The trend in the graph reveals that the occurrence of central explanatory sentences tends to follow a long tail distribution, with a small set of sentences frequently reused across different explanations. This trend suggests that a subsets of sentences results particularly useful to construct explanations for many science questions, constituting a first indication that some central sentence might possess a greater \emph{explanatory power} and induce certain \emph{patterns of unification}.

\begin{table*}[t]
\centering
\small
\begin{tabular}{p{14cm}c}
\toprule
\textbf{Central Explanatory Sentence} & \textbf{Occurrence}\\
\midrule
Boiling;evaporation means matter; a substance changes from a liquid into a gas by increasing heat energy & 33\\
\midrule
An adaptation; an ability has a positive impact on an animal's; living thing's survival; health; ability to reproduce & 24\\
\midrule
Photosynthesis means producers; green plants convert from carbon dioxide and water and solar energy into carbohydrates and food and oxygen for themselves & 23 \\
\midrule
Inheriting is when an inherited characteristic is copied; is passed from parent to offspring by genetics; DNA & 23 \\
\midrule
Melting means matter; a substance changes from a solid into a liquid by increasing heat energy & 21 \\
\midrule
If an object is made of a material then that object has the properties of that material & 20\\
\midrule
Photosynthesis is a source of; makes food; energy for the plant by converting carbon dioxide, water, and sunlight into carbohydrates & 20\\ 
\midrule
Water is in the solid state , called ice , for temperatures between 0; -459; -273 and 273; 32; 0 K; F; C & 16 \\ 
\midrule
Decomposition is when a decomposer breaks down dead organisms 16
an animal; living thing requires nutrients for survival & 16 \\
\midrule
Objects are made of materials; substances; matter & 15 \\
\midrule
Chemical reactions cause new substances; different substances to form & 15 \\
\bottomrule
\end{tabular}
\caption{Most reused central explanatory sentences in WorldTree.}
\label{tab:central_facts_example}
\end{table*}

\begin{table*}[t]
\centering
\small
\ra{1}
\resizebox{\textwidth}{!}{
\begin{tabular}{p{5cm}p{7cm}c}
\toprule
\textbf{Grounding} & \textbf{Grounding} & \textbf{Occurrence}\\
\midrule
\_ is a kind of \_ (Taxonomic) & \_ is a kind of \_ (Taxonomic) & 524\\
\_ is a kind of \_ (Taxonomic) & \_ is part of \_ (Part-of) & 73\\
\_ is a kind of \_ (Taxonomic) & \_ is made of \_ (Made-of) & 37\\
\_ is a kind of \_ (Taxonomic) & \_ typically performs action \_ on \_ (Actions) & 30\\
\_ is a kind of \_ (Taxonomic) & \_ is a property of \_ (Properties) & 25\\
\bottomrule
\toprule
\textbf{Grounding} & \textbf{Central} & \textbf{Occurrence}\\
\midrule
\_ is a kind of \_ (Taxonomic) & \_ typically performs action \_ on \_ (Actions) & 209\\
\_ is a kind of \_ (Taxonomic) & if \_ then \_ (Conditionals) & 202\\
\_ is a kind of \_ (Taxonomic) & \_ causes \_ (Causal) & 179\\
\_ is a kind of \_ (Taxonomic) & \_ changes from \_ to \_ by \_ (Processses) & 153\\
\_ is a kind of \_ (Taxonomic) & \_ uses \_ for \_ (Functional) & 133\\
\bottomrule
\end{tabular}}
\caption{Most reused categories of grounding-central inference chains in WorldTree.}
\label{tab:grounding_central_chains_knowledge}
\end{table*}

\begin{table*}[t]
\centering
\small
\ra{1}
\resizebox{\textwidth}{!}{
\begin{tabular}{p{6cm}p{6cm}c}
\toprule
\textbf{Grounding} & \textbf{Grounding} & \textbf{Occurrence}\\
\midrule
An animal is a kind of living thing & A living thing is a kind of object & 18\\
\midrule
An animal is a kind of organism & A plant is a kind of organism & 14\\
\midrule
A human is a kind of animal & An animal is a kind of organism & 14\\
\midrule
A tree is a kind of plant & A plant is a kind of organism & 11\\
\midrule
A human is a kind of animal & An animal is a kind of living thing & 11\\
\bottomrule
\toprule
\textbf{Grounding} & \textbf{Central} & \textbf{Occurrence}\\
\midrule
Water is a kind of liquid at room temperature &   Boiling;evaporation means matter; a substance changes from a liquid into a gas by increasing heat energy & 20\\
\midrule
Metal is a kind of material & If an object is made of a material then that object has the properties of that material & 14\\
\midrule
Earth is a kind of planet & A planet rotating causes cycles of day and night on that planet & 9\\
\midrule
A plant is a kind of organism & Decomposition is when a decomposer breaks down dead organisms & 9\\
\midrule
Water is a kind of liquid at room temperature & Freezing means matter; a substance changes from a liquid into a solid by decreasing heat energy & 9\\
\midrule 
Metal is a kind of material & Metal is a thermal; thermal energy conductor & 9\\
\bottomrule
\end{tabular}}
\caption{Most reused sentence-level inference chains in WorldTree.}
\label{tab:grounding_central_chains_sentence}
\end{table*}

To further investigate this aspect, Figure \ref{fig:central_occurrence_similarity} correlates the frequencies of central explanatory sentences in the corpus ($x$ axis) with the average similarity between the same sentences and the questions they explain ($y$ axis). To perform the analysis, the similarity values are computed adopting BM25 and cosine distance between each question and its explanation sentences \cite{robertson2009probabilistic}. From a unificationist point of view, we expect to find an inverse correlation between the frequency of reuse of a central sentence and its similarity with the explanandum. Specifically, we assume that the lower the similarity, the higher the probability that a central sentence describes abstract laws and high level regularities, and that, therefore, it is able to \emph{unify} a larger set of phenomena. Under this assuptions and considering naturally occurring variability in the dataset, the trend in Figure \ref{fig:central_occurrence_similarity} confirms the expectation, showing that the most reused central sentences are also the one that explain clusters of less similar questions. In particular, the graph reinforces the hypothesis that the reuse value of a central sentence in the corpus is indeed connected with its \emph{unification power}.

The concrete examples in Table \ref{tab:central_facts_example} further support this hypothesis. Specifically, the table shows that it is possible to draw a parallel between the distribution of central sentences in the corpus and the notion of \emph{argument patterns} in the Unificationist account \cite{kitcher1981explanatory}. It is possible to notice, in fact, that the most occurring central sentences tend to describe high-level processes and regularities, typically mentioning abstract concepts and entities (e.g., \emph{living things}, \emph{object}, \emph{substance}, \emph{material}). In particular, the examples suggest that reoccurring central explanatory facts might act as \emph{schematic sentences} of an \emph{argument pattern}, with abstract entities representing the linguistic counterpart of \emph{variables} and \emph{filling instructions} used to specify and constraining the space of possible instantiations.

\subsubsection{Abstraction and Patterns of Unification}



To further explore the parallel between natural language explanations and the Unificationist account, we focus on recurring inference chains between \emph{grounding} and \emph{central} sentences. Specifically, we aim at investigating whether it is possible to map inference patterns in WorldTree to the process of instantiating \emph{schematic sentences} for unification. To this end, we automatically build a linkage between grounding and central sentences in the corpus using the support of lexical overlaps.

Table \ref{tab:grounding_central_chains_knowledge} reports the most recurring linguistic categories of grounding-central chains, which provide and indication of the high-level process through which explanatory patterns emerge in natural language. Overall, we found a clear evidence of inference patterns related to the \emph{instantiation} of central explanatory sentences. Specifically, the table shows that these patterns emerge through the use of taxonomic knowledge. This suggests that abstraction, intended as the process of going from concrete concepts in the explanandum to high-level concepts in the explanans, is a fundamental part of the inference required for explanation and it is what allows submsuming the explanandum under unifying regularities. 
Central sentences, in fact, tend to be represented by a more diverse set of linguistic categories in line with those described in the philosophical accounts (i.e., causes, processes, general rules). 
By looking at grounding-grounding connections, it is possible to notice the relatively high frequency of chains of taxonomic relations, which confirms again the parallel between explanatory patterns in the corpus and the process of instantiating abstract schematic sentences for unification. Moreover, the presence of linguistic elements about generic attributes and properties is in line with the analysis on the Biology Why Corpus, supporting the fact that these pragmatic elements in natural language explanations play an important role in the abstraction-instantiation process.
Table \ref{tab:grounding_central_chains_sentence} shows examples of sentence-level explanatory patterns, demonstrating how the process of abstraction and unification concretely manifests in the corpus.

Overall, it is possible to conclude that explanatory patterns emerging in natural language explanations are closely related to unification, and that this process is fundamentally supported by an inference substrate performing abstraction, whose function is to connect the explanandum to the description of high-level patterns and unifying regularities. 

\subsection{Summary}

\begin{figure}[t]
\centering
\includegraphics[width=\textwidth]{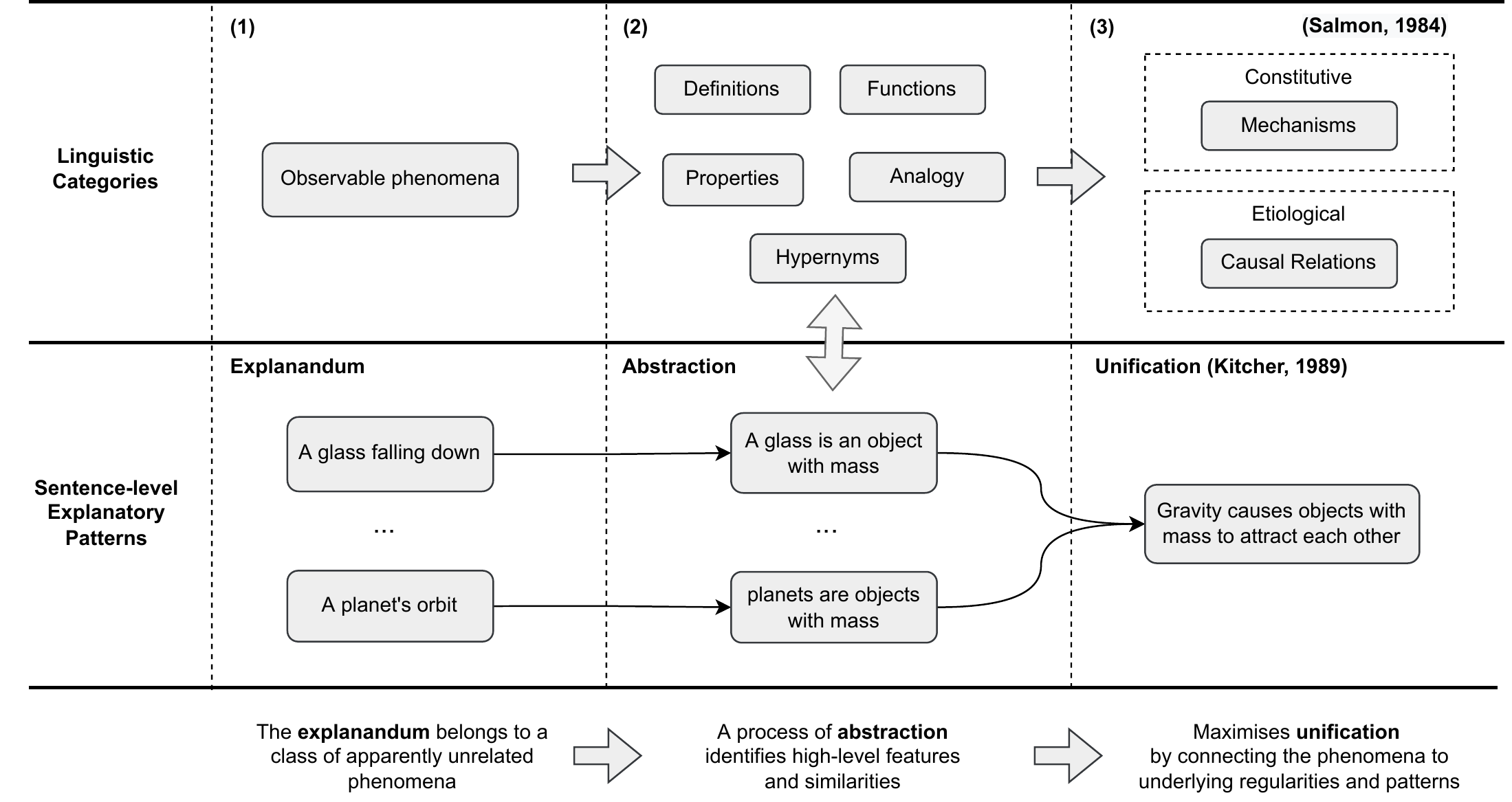}
\caption{A synthesis between the formal accounts of scientific explanations and linguistic aspects found through the corpus analysis.}
\label{fig:explanation_synthesis}
\end{figure}

The main results and findings of the corpus analysis can be summarised as follows:

\begin{enumerate}
    \item \textbf{Natural language explanations are not limited to causes and mechanisms.} While \emph{constitutive} and \emph{etiological} elements represent the core part of an explanation, our analysis suggests that additional knowledge categories such as \emph{definitions}, \emph{properties} and \emph{taxonomic relations} play an equally important role in natural language. This can be attributed to both \emph{pragmatic aspects} of explanations and inference requirements associated to \emph{unification}.
    \item \textbf{Patterns of unification naturally emerge in corpora of explanations.} Even if not intentionally modelled,  \emph{unification} seems to be an emergent property of corpora of natural language explanations. The corpus analysis, in fact, reveals that the frequency of reuse of certain explanatory sentences is connected with the notion of \emph{unification power}. Moreover, a qualitative analysis suggests that reused explanatory facts might act as \emph{schematic sentences}, with abstract entities representing the linguistic counterpart of \emph{variables} and \emph{filling instructions} in the Unificationist account.
    \item \textbf{Unification is realised through a process of abstraction.} Specifically, abstraction represents the fundamental inference substrate supporting unification in natural language. The corpus analysis, in fact, suggests that recurring explanatory patterns emerge through inference chains connecting concrete instances in the explanandum to high-level concepts in the central explanans. This process, realised through specific linguistic elements such as \emph{definitions} and \emph{taxonomic relations}, is a fundamental part of natural language explanations, and represents what allows subsuming the event to be explained under high-level patterns and unifying regularities.
\end{enumerate}

\section{Synthesis}

Finally, with the help of Figure \ref{fig:explanation_synthesis}, it is possible to perform a synthesis between the epistemological accounts of scientific explanation and the linguistic aspects emerging from the corpus analysis.

In general, explanations cannot be exclusively characterised in terms of \emph{inductive} or \emph{deductive} arguments. This because the logical structure of explanations and predictions is intrinsically different \cite{woodward2003scientific}. From an epistemic perspective, in fact, the main function of an explanatory argument is to fit the explanandum into a broader pattern that maximises unification, showing that a set of apparently unrelated phenomena are part of a common regularity \cite{kitcher1981explanatory,kitcher1989explanatory}. From a linguistic point of view, the process of unification tends to generate sentence-level \emph{explanatory patterns} that can be reused and instantiated for deriving and explaining many phenomena.  In natural language, unification generally emerges as a process of \emph{abstraction} from the explanandum through the implicit search of common high-level features and similarities between different phenomena.

From an ontic perspective, causal interactions and mechanisms constitute the central part of an explanation as they make the difference between the occurrence and non occurrence of the explanandum \cite{salmon1984scientific,lipton1990contrastive}. Moreover, causal interactions are responsible for high-level regularities and invariants, with many phenomena being the result of the same underlying causal mechanisms. Here, abstraction represents the inference substrate linking the explanandum to these regularities, a process that manifests in natural language through the use of specific linguistic elements coupled with causes and mechanisms, such as definitions, taxonomic relations, and analogies.



\section{Implications for Explanation-based AI}

Current lines of research in Explainable AI focus on the development and evaluation of explanation-based models, capable of performing inference through the generation of natural language explanations \cite{wiegreffe2021teach,xie2020worldtree,jansen2018worldtree,thayaparan-etal-2021-textgraphs}. 

Evaluating quality and properties of natural language explanations is still extremely challenging \cite{jansen2021challenges,valentino2021natural}, with most of the existing work focusing on inferential properties in terms of \emph{entailment} or \emph{supporting facts} \cite{yang2018hotpotqa,camburu2018snli,valentino2021natural,dalvi2021explaining}. This study, however, shows that natural language explanations cannot be evaluated exclusively in terms of deductive reasoning and entailment. This because deductive arguments cannot fully characterise explanations, and cannot distinguish explanatory arguments from mere predictive ones. As the main function of an explanation is to perform unification, the evaluation methodologies should explicitly reflect this property. 

Regarding the construction of explanation-centred corpora, while unification seems to be an emergent property of existing datasets \cite{xie2020worldtree,jansen2018worldtree}, future research can benefit from explicitly considering it during the annotation process. Unification patterns, in fact, can provide a top-down and reuse-oriented methodology to facilitate evaluation and scale up the annotation process.
From an inferential perspective, the evaluation of natural language explanations should focus on a multi-dimensional set of inference capabilities, assessing explanation-based systems in the ability to perform abstraction, identify unifying causal mechanisms and interpret high-level regularities.

Finally, emergent unification patterns in natural language explanations can provide a way to build more robust inference models \cite{valentino2021unification}. Recent work in multi-hop natural language inference, in fact, has shown that recurring explanatory patterns can potentially reduce the search space for the automatic generation of natural language explanations \cite{valentino2021hybrid,valentino2020case} and provide support for a better design of abstractive inference capabilities \cite{thayaparan-etal-2021-explainable}. 

\section{Conclusion}

In order to provide an epistemologically grounded characterisation of natural language explanations, this paper attempted to bridge the gap in the notion of \emph{scientific explanation} \cite{salmon2006four,salmon1984scientific}, studying it as both a \emph{formal object} and as a \emph{linguistic expression}. The combination of a systematic survey with a corpus analysis on natural language explanations \cite{jansen2014discourse,jansen2018worldtree}, allowed us to derive specific conclusions on the nature of explanatory arguments from both a top-down (categorical) and a bottom-up (corpus-based) perspective:
\begin{enumerate}
    \item Explanations cannot be entirely characterised in terms of \emph{inductive} or \emph{deductive} arguments as their main function is to perform \emph{unification}.
    \item A scientific explanation must cite causes and mechanisms that are responsible for the occurrence of the explanandum.
    \item While natural language explanations possess an intrinsic causal-mechanistic nature, they are not limited to causes and mechanisms.
    \item Patterns of unification naturally emerge in corpora of explanations even if not intentionally modelled.
    \item Unification emerges through a process of abstraction, whose function is to provide the inference support for subsuming the event to be explained under recurring patterns and regularities.
\end{enumerate}

From these findings, it is possible to derive a set of guidilines for furure research on Explainable AI for the creation and evaluation of models that can interpret and generate natural language explanations:
\begin{enumerate}
    \item Explainability cannot be evaluated only in terms of deductive inference capabilities and entailment properties. This because deductive arguments cannot entirely characterise explanations, and cannot be used to distinguish explanatory arguments from mere predictive ones.
    \item As the main function of an explanatory arguments is to perform unification, the evaluation of explainability must explicitly take into account this property. Moreover, while unification seems to be an emergent property of existing benchmarks, it should be explicitly considered as a top-down approach for the creation of explanation-centred corpora to facilitate evaluation.
    \item From a bottom-up perspective, the evaluation of explainability should not only focus on specific inference properties connected to causality, but also take into account other features of explanation, including semantic abstraction and analogy.
    \item The unification property of explanatory arguments can provide a way to build more robust inference models that explicitly leverage patterns of derivation, as well as more efficient and scalable solutions to construct explanation-centred corpora. Recurring argument patterns, in fact, can potentially reduce the search space for multi-hop inference models and support a more schematic, reuse-oriented mechanism for the annotation of gold explanations.
\end{enumerate}

The paper contributed to addressing a fundamental gap in classical theoretical accounts on the nature of scientific explanations and their materialisation as linguistic artefacts, providing a unified epistemological-linguistic perspective. This characterisation can support a more principled design and evaluation of explanation-based AI systems which can better interpret and generate natural language explanations.

\bibliographystyle{coling}
\bibliography{references}

\end{document}